\documentclass{article}

\usepackage{microtype}
\usepackage{graphicx}
\usepackage{booktabs} 
\usepackage{xspace}

\usepackage{amssymb}
\usepackage{amsmath}
\usepackage[frozencache,cachedir=.]{minted}
\AtBeginEnvironment{minted}{\footnotesize}
\usepackage{multirow}
\usepackage{subcaption}
\usepackage{hyperref}
\usepackage{tikz}
\usetikzlibrary{arrows.meta, positioning, fit, calc, backgrounds}

\usepackage[accepted]{mlsys2025-arxiv}

\def\Snospace~{\S{}}

\newcommand{\sys}{AXLearn\xspace}
\newcommand{\paraspace}{\vspace{0.05in}}
\newcommand{\parab}[1]{\paraspace\noindent{\bf #1} }

\mlsystitlerunning{\sys: Modular, Hardware-Agnostic Large Model Training}

\begin{document}

\twocolumn[
\mlsystitle{\sys: Modular, Hardware-Agnostic Large Model Training}



\mlsyssetsymbol{equal}{*}

\begin{mlsysauthorlist}
\mlsysauthor{Mark Lee$^*$}{apple}
\mlsysauthor{Chang Lan$^*$}{apple}
\mlsysauthor{Tom Gunter$^\dagger$}{apple}
\mlsysauthor{John Peebles$^\dagger$}{apple}
\mlsysauthor{Hanzhi Zhou$^\dagger$}{apple}
\mlsysauthor{Kelvin Zou$^\dagger$}{apple}

\mlsysauthor{Sneha Bangalore}{apple}
\mlsysauthor{Chung-Cheng Chiu}{apple}
\mlsysauthor{Nan Du}{apple}
\mlsysauthor{Xianzhi Du}{apple}
\mlsysauthor{Philipp Dufter}{apple}
\mlsysauthor{Liang He}{apple}
\mlsysauthor{Ruixuan Hou}{apple}
\mlsysauthor{Haoshuo Huang}{apple}
\mlsysauthor{Dongseong Hwang}{apple}
\mlsysauthor{Xiang Kong}{apple}
\mlsysauthor{Jinhao Lei}{apple}
\mlsysauthor{Tao Lei}{apple}
\mlsysauthor{Meng Li}{apple}
\mlsysauthor{Li Li}{apple}
\mlsysauthor{Jiarui Lu}{apple}
\mlsysauthor{Zhiyun Lu}{apple}
\mlsysauthor{Yiping Ma}{apple}
\mlsysauthor{David Qiu}{apple}
\mlsysauthor{Vivek Rathod}{apple}
\mlsysauthor{Senyu Tong}{apple}
\mlsysauthor{Zhucheng Tu}{apple}
\mlsysauthor{Chong Wang}{apple}
\mlsysauthor{Jianyu Wang}{apple}
\mlsysauthor{Yongqiang Wang}{apple}
\mlsysauthor{Zirui Wang}{apple}
\mlsysauthor{Floris Weers}{apple}
\mlsysauthor{Sam Wiseman}{apple}
\mlsysauthor{Guoli Yin}{apple}
\mlsysauthor{Bowen Zhang}{apple}
\mlsysauthor{Xiyou Zhou}{apple}
\mlsysauthor{Danyang Zhuo$^\S$}{duke}
\mlsysauthor{Cheng Leong}{apple}
\mlsysauthor{Ruoming Pang$^\ddag$}{apple}
\end{mlsysauthorlist}

\mlsysaffiliation{apple}{Apple}
\mlsysaffiliation{duke}{Duke University}

\additionalinfo{$^*$ Mark Lee and Chang Lan are the first authors. $^\dagger$ denotes the core authors. $^\S$Danyang Zhuo contributed to this work as a visiting scholar at Apple. $^\ddag$ Ruoming Pang is the corresponding author.}


\vskip 0.3in

\begin{abstract}
\sys is a production system which facilitates scalable and high-performance training of large deep learning models. Compared to other state-of-art deep learning systems, \sys has a unique focus on modularity and support for hardware-agnostic training. \sys's internal interfaces between software components follow strict encapsulation, allowing different components to be assembled to facilitate rapid model development and experimentation on different hardware infrastructure.
\sys maintains constant complexity as we scale the components in the system, compared to linear or quadratic complexity in state-of-the-art training systems. This allows integrating features such as Rotary Position Embeddings (RoPE) into \sys across hundred of modules with just 10 lines of code, compared to hundreds as required in other systems.
At the same time, \sys maintains equivalent performance compared to state-of-the-art training systems. Finally, we share our experience in the development and operation of \sys at Apple.
\end{abstract}
]



\printAffiliationsAndNotice{}  

\section{Introduction}

Large-scale deep learning models are now integral to society—they power chatbots such as ChatGPT and Gemini~\cite{chatgpt, gemini}, enhance video conferencing~\cite{zoom}, and support modern coding tools~\cite{cursor}. 
Modern deep learning systems prioritize performance and scalability to accommodate large models. Numerous techniques have been explored to this end, including parallelization strategies~\cite{narayanan2021megatron, zheng2022alpa}, memory optimizations~\cite{Rajbhandari2019zero, zhao2023pytorchfsdp}, and model-specific kernel optimizations~\cite{dao2022flashattention, dao2023flashattention2}.

At Apple, we have integrated many AI models into our products, catering to billions of users worldwide.
We have two requirements for our deep learning systems besides training performance and scalability. First, we aim to empower our model engineers to experiment with diverse model architectures and training techniques. They should write only a minimal amount of code to configure complex model definitions and training methods. We call this \textit{modularity} of the deep learning system.
Second, as a large technology company, we cannot rely on a single hardware vendor---any hardware can run into supply issues and vary in pricing. Our design goal is be \textit{hardware-agnostic}, supporting GPU, TPU, and AWS Trainium. This allows us to train on major cloud providers (e.g., AWS, Google Cloud, Azure) as well as our on-premises servers.

\begin{table*}[t]
    \caption{Comparing \sys with state-of-the-art deep learning systems. We define ``partial" as systems that decouple parallelism from model implementation via device abstractions (e.g., XLA), but in practice do not apply strict encapsulation.}
    \begin{center}
	\begin{small}
		\begin{tabular}{@{}c|c|c|c|c|c|c|c@{}}
		\toprule
        System & Underlying Framework & Model Agnostic & 3D Parallelism & Modular & GPU & TPU & Trainium \\
	    \hline
        Megatron-LM & PyTorch &  & $\checkmark$ & & $\checkmark$ &  \\
        \hline
        DeepSpeed & PyTorch & $\checkmark$ & $\checkmark$ & & $\checkmark$ &   \\
        \hline
        PyTorch FSDP & PyTorch & $\checkmark$ &  & & $\checkmark$ & \\
        \hline
        PyTorch XLA FSDP & PyTorch & $\checkmark$ &  & & $\checkmark$ & $\checkmark$ \\
        \hline
        TorchTitan & PyTorch &  & $\checkmark$ & partial & $\checkmark$ &   \\
        \hline
        Haiku & JAX & $\checkmark$ & $\checkmark$ & partial & $\checkmark$ & $\checkmark$ &  \\
        \hline
        Flax & JAX & $\checkmark$ & $\checkmark$ & partial & $\checkmark$ & $\checkmark$\\
        \hline
        Pax & JAX & $\checkmark$ & $\checkmark$ & partial & $\checkmark$ &  $\checkmark$\\
        \hline
        MaxText & JAX & &$\checkmark$ & partial & $\checkmark$ & $\checkmark$ & \\
        \hline
		\sys (Ours) & JAX & $\checkmark$ & $\checkmark$ & $\checkmark$ & $\checkmark$ & $\checkmark$ & $\checkmark$ \\
		\bottomrule
		\end{tabular}
	\end{small}
    \end{center}
    \label{tab:mlsys}
\end{table*}  

To facilitate modularity, the core design decision of \sys is to enforce \textit{strict encapsulation}. 
While encapsulation is a well known principle in object oriented programming, we find that it's often neglected in ML frameworks as almost all existing frameworks rely on subtyping. 
The prevalence of subtyping can be partially attributed to the lack of formal analysis on how to \textit{quantify} the modularity and extensibility of a system, where conventional design principles or ``rules of thumb" are non-exhaustive and hard to measure. 
To this end, we propose a framework for quantifying the complexity of a system by measuring \textit{asymptotic} LoC changes incurred by the addition of a new feature.
We also show that the framework is consistent with more traditional ways of counting LoC via the case study of integrating Rotary Positional Embeddings (RoPE)~\cite{su2023roformerenhancedtransformerrotary} and Mixture of Experts (MoE)~\cite{shazeer2017outrageouslylargeneuralnetworks} into \sys as compared to other state-of-the-art deep learning systems.
To our knowledge, \sys is the only training framework that adheres strictly to encapsulation---any module is replaceable, including the input pipeline, checkpointer, trainer loop---allowing complex features to be implemented without increasing the complexity of the overall system.

To enable hardware-agnostic training, we build \sys on top of XLA~\cite{xla} and GSPMD~\cite{xu2021gspmd}.
Our native integration with XLA allows parallelism strategies to be automatically generated, but still allows hand-crafting kernel code for specific accelerators. For instance, on each hardware backend, we replace the attention layer with a custom kernel like FlashAttention~\cite{dao2022flashattention, dao2023flashattention2}. 
We believe this design strikes an ideal balance: by leveraging the XLA ecosystem, we can seamlessly support multiple hardware accelerators without sacrificing high performance. 
Due to its modular design, \sys also enables succinct user configurations to customize the parallelization, rematerialization, and quantization strategies, further simplifying the scaling experience.

\autoref{tab:mlsys} provides a list of deep learning systems designed for large models. 
Besides modularity and hardware requirement, the table indicates whether each system is model-agnostic or supports broader parallelism strategies---such as 3D parallelism~\cite{narayanan2021megatron}---to enable efficient training across diverse model architectures. 
We summarize several important observations. Some systems only support specific architectures: 
for example, Maxtext~\cite{maxtext} provides LLM implementations but does not extend easily to custom architectures as its design encourages fork-and-modify rather than reusing building blocks.
Similarly, TorchTitan~\cite{liang2024torchtitan} requires model-specific parallel plans that traverse the model architecture, limiting its applicability to architectures not explicitly supported.
Other systems are designed for specific backends: for instance, Megatron-LM~\cite{narayanan2021megatron} has carefully designed optimizations for transformers on GPU, but these optimizations do not directly apply to other hardware.
Apple uses a diverse set of model architectures across a range of cloud backends, so we cannot directly use these systems.

Over the past few years, we've deployed \sys to train thousands of models involving hundreds of engineers.
The rapid adoption of \sys largely owes to its modularity and unique ability to scale on various public clouds, including Google Cloud TPU, AWS GPU or Trainium2. 
We share how our experience evolved our design choices over several years. 
\sys is open-sourced under Apache 2.0 license at \url{https://github.com/apple/axlearn}.

\section{Motivation}
\label{sec:background}

\subsection{Modularity}

In existing ML systems, neural network layers are implemented by subtyping: a layer inherits from some base layer, instantiates child layers as instance attributes, and implements a \texttt{forward} method that handles the layer's logic. Consider the changes required to replace a feed-forward network (FFN) of a Transformer architecture with an MoE layer using subtyping. Taking the example from the DeepSpeed~\cite{deepspeed_moe_tutorial}, one applies such a change by replacing the instance attribute for the original FFN:

\begin{minted}{python}
- self.fc3 = nn.Linear(84, 10)
+ self.fc3 = nn.Linear(84, 84)
+ self.fc3 = deepspeed.moe.layer.MoE(...)
+ self.fc4 = nn.Linear(84, 10)
\end{minted}

At a glance, this seems like a simple 4 LoC change. However, in practice one would need to subtype \text{self.fc3}'s parent layer to apply such a change. This effectively reduces the problem of replacing the FFN with MoE, to the problem of replacing the parent layer with the new subtyped layer. By induction, it's easy to see how such a change can compound to changes to multiple modules across the subtype hierarchy. 
Indeed, this happens in practice: between DeepSpeed's QwenV2 and QwenV2 MoE implementations~\cite{deepspeed_models}
over 200 LoC are required to apply the MoE layer, not accounting for the MoE layer itself---a far cry from the 4 LoC that we may have hoped.

Conceptually, MoE is a small change---ideally, one should be able to ``swap'' the FFN directly with an MoE equivalent, without incurring the side-effects of propagating changes up to ancestors like the Transformer layer.
In \sys, this can be achieved by exploiting the compositional nature of neural networks: by implementing the FFN and MoE layers with compatible input/output interfaces, and \textit{encapsulating} all MoE specific details within its layer implementation, we realize this ideal scenario of treating MoE as a ``drop-in'' replacement.

Note that by naively comparing the LoC between implementations, the benefits of composition cannot be fully observed.
Specifically, the 190 LoC difference between the two implementations does not account for higher-order effects: if we consider that a production codebase may instead contain tens to hundreds of variants of the same model, the 200 LoC change quickly becomes thousands. 


\begin{figure}[t]
\centering
\resizebox{0.95\linewidth}{!}{%
\begin{tikzpicture}[
  every node/.style={font=\normalsize\sffamily, inner sep=5pt, minimum height=0.5cm, minimum width=1.8cm},
  reused/.style={fill=red!8, draw=red!40, rounded corners=3pt, line width=0.8pt},
  new/.style={fill=green!10, draw=green!50!black!40, rounded corners=3pt, line width=0.8pt},
  leaf/.style={reused, minimum width=1.8cm, font=\footnotesize\sffamily},
  edge/.style={draw=gray!70, line width=0.8pt},
  dots/.style={font=\normalsize\sffamily, text=gray!60, inner sep=2pt, minimum width=0cm},
]
\def\vs{0.4cm}   
\def\mg{0.35cm}   

\node[reused] (a-trans) at (1.0,0) {Transformer};
\node[reused] (a-sa)  at (5, 0) {Self Attn};
\node[reused] (a-ffn) at (5,-2.4) {FFN};
\node[reused] (a-attn) at (9, 0) {Attn};
\node[reused] (a-qkv)  at (9, -0.8) {QKV proj};
\node[reused]   (a-dots) at (9, -1.6) {\ldots};
\node[reused] (a-l1) at (9,-2.4) {Linear 1};
\node[reused] (a-l2) at (9,-3.2) {Linear 2};
\coordinate (a-m0) at ($(a-trans.east)+(\mg,0)$);
\draw[edge] (a-trans.east) -- (a-m0);
\draw[edge] (a-m0) |- (a-sa.west);
\draw[edge] (a-m0) |- (a-ffn.west);
\coordinate (a-m1) at ($(a-sa.east)+(\mg,0)$);
\draw[edge] (a-sa.east) -- (a-m1);
\draw[edge] (a-m1) |- (a-attn.west);
\draw[edge] (a-m1) |- (a-qkv.west);
\draw[edge] (a-m1) |- (a-dots.west);
\coordinate (a-m2) at ($(a-ffn.east)+(\mg,0)$);
\draw[edge] (a-ffn.east) -- (a-m2);
\draw[edge] (a-m2) |- (a-l1.west);
\draw[edge] (a-m2) |- (a-l2.west);
\node[below=0.6cm of a-ffn, font=\normalsize, minimum width=0cm] {(a) Standard transformer};

\def\by{-4.8}  
\node[reused] (b-trans) at (0,\by) {Transformer};
\node[reused] (b-sa)  at (2.5, \by) {Self Attn};
\node[reused] (b-ffn) at (2.5, \by-2.4) {FFN};
\node[reused] (b-attn) at (5.0, \by) {Attn};
\node[reused] (b-qkv)  at (5.0, \by-0.8) {QKV proj};
\node[reused]   (b-dots) at (5.0, \by-1.6) {\ldots};
\node[new]    (b-moe) at (5.0, \by-2.4) {MoE};
\node[new]  (b-gate) at (7.5, \by) {Gating};
\node[reused]  (b-ffn1) at (7.5, \by-0.8) {FFN 1};
\node[reused] (b-dots2) at (7.5, \by-1.6) {\ldots};
\node[reused]  (b-ffnk) at (7.5, \by-2.4) {FFN K};
\node[leaf] (b-f1-l1) at (10, \by) {Linear 1};
\node[leaf] (b-f1-l2) at (10, \by-0.8) {Linear 2};
\node[leaf] (b-fk-l1) at (10, \by-1.6) {Linear 1};
\node[leaf] (b-fk-l2) at (10, \by-2.4) {Linear 2};
\coordinate (b-m0) at ($(b-trans.east)+(\mg,0)$);
\draw[edge] (b-trans.east) -- (b-m0);
\draw[edge] (b-m0) |- (b-sa.west);
\draw[edge] (b-m0) |- (b-ffn.west);
\coordinate (b-m1) at ($(b-sa.east)+(\mg,0)$);
\draw[edge] (b-sa.east) -- (b-m1);
\draw[edge] (b-m1) |- (b-attn.west);
\draw[edge] (b-m1) |- (b-qkv.west);
\draw[edge] (b-m1) |- (b-dots.west);
\draw[edge] (b-ffn.east) -- (b-moe.west);
\coordinate (b-m3) at ($(b-moe.east)+(\mg,0)$);
\draw[edge] (b-moe.east) -- (b-m3);
\draw[edge] (b-m3) |- (b-gate.west);
\draw[edge] (b-m3) |- (b-ffn1.west);
\draw[edge] (b-m3) |- (b-dots2.west);
\draw[edge] (b-m3) |- (b-ffnk.west);
\coordinate (b-m4) at ($(b-ffn1.east)+(\mg,0)$);
\draw[edge] (b-ffn1.east) -- (b-m4);
\draw[edge] (b-m4) |- (b-f1-l1.west);
\draw[edge] (b-m4) |- (b-f1-l2.west);
\coordinate (b-m5) at ($(b-ffnk.east)+(\mg,0)$);
\draw[edge] (b-ffnk.east) -- (b-m5);
\draw[edge] (b-m5) |- (b-fk-l1.west);
\draw[edge] (b-m5) |- (b-fk-l2.west);
\node[below=0.3cm of b-moe, font=\normalsize, minimum width=0cm] {(b) MoE transformer};
\end{tikzpicture}%
}
\caption{Specifying MoE transformer in \sys. Red components are reused from the specification of standard transformer. In \sys, a user script that defines MoE only needs to specify the green parts of the neural network.}
\label{fig:moe}
\end{figure}

We instead propose to measure the extensibility of a system in terms of the \textit{asymptotic} LoC changes required by some re-parameterization of its API. 
When we refer to the re-parameterization of the API, we essentially pose the question of how the LoC changes when we reconfigure the system to support a different implementation signature---such as adding new functionality like the MoE layer.
This metric can be used to explain why composition should be favored over subtyping. The LoC-complexity(MoE) for DeepSpeed is lower-bounded by $O(N)$, where $N$ denotes the number of modules in the system (\autoref{sec:evaluation}): we require (1) at least one LoC change for each attention variant to subtype a \texttt{forward} implementation that replaces FFN with the MoE equivalent; (2) at least one LoC change in each ancestor module of an attention layer to incorporate the new subtyped layer. 
In contrast, the LoC-complexity(MoE) in \sys is $O(1)$: the code snippet in \autoref{sec:compiler} can be used to integrate MoE without changing \textit{any} model.

In practice, we can validate the LoC-complexity with concrete LoC counts. Indeed, in \sys we use the same 10-line snippet to configure MoE in over 1,000 different experiments\footnote{An experiment refers to a training job configuration, such as the model architecture and hyper-parameters.}.  
On the other hand, if we were to adopt DeepSpeed's strategy of integrating MoE to our internal codebase, we would incur over 4,000 LoC just to modify different model variants to support their MoE counterparts.


\subsection{Hardware-Agnostic Training}

As one of the largest technology companies, Apple cannot feasibly rely on a single hardware platform for all of our machine learning workloads. 
Aside from supply constraints, our company is positioned to benefit from \textit{not} committing to any single platform: Megatron-LM~\cite{narayanan2021megatron} has a vested interest in optimizing for Nvidia GPUs, while Haiku~\cite{haiku}, Flax~\cite{flax}, Pax~\cite{paxml}, and MaxText~\cite{maxtext} are mostly optimized for Google TPUs.

Instead, we take the alternate approach of developing a hardware-agnostic system that can adapt seamlessly to multiple platforms, which allows our engineers to take advantage of the most cost-effective solution for their training workload. For example, while AWS Trainium2 did not exist when we first began development, \sys is one of the first deep learning system that supports Trainium2 at scale.

While JAX/XLA is a key component to support hardware-agnostic training, it is not sufficient. XLA often provides reasonable out-of-the-box performance, but additional steps are needed to achieve the best performance---including providing ``hints'' to the XLA compiler at various points in the layer graph, using hand-tuned kernels for less mature compilation targets like GPU or Trainium2, or customizing the rematerialization strategy for each workload. 

\section{Overview}
\label{sec:overview}

\sys allows model engineers to rapidly experiment with various model architecture and training methods across different hardware backends.
\sys's developer interface is a hierarchical configuration of \textit{modules}. A module can be viewed abstractly as a node in an object tree. In training, the root module is typically the trainer itself, with child modules such as the model, learner, and input, each of which may have its own children. A module's definition follows a consistent structure, including a config object that encapsulates all configurable parameters of the module. 

Our view is that for ML researchers, writing configs and composing modules is much easier than implementing subtyped modules from existing ones, a view that is shared by TorchTitan~\cite{liang2024torchtitan}. The pure composition approach allows replacing individual components and makes it easy to reuse components between teams or 3rd party libraries.
Similar to TorchTitan, our model definition is independent of any specific parallelization strategy or trainer loop.
However, from the developer experience point of view, \sys has several key differences in how modules are parameterized. 
First, \sys encourages hierarchical config composition over config flattening. In the example shown in \autoref{fig:moe}, a user script can build on top of a standard transformer architecture by defining only the green components, and selectively replacing FFN modules with MoE equivalents without needing to inspect the details of the base transformer architecture.
In contrast, TorchTitan adopts a monolithic approach where config files ``flatten'' all possible configurations, including model, optimizer, checkpointer, and trainer configs. Flattening has significant ramifications in the extensibility of TorchTitan. (See \autoref{sec:evaluation}.)
Second, \sys uses an entirely Python-based interface, which has the benefit of allowing configs to be expressed with Python constructs like functions, loops, and recursion; as well as the advantage of being able to be directly unit-tested. 

\begin{figure}
\centering
\includegraphics[width=0.95\linewidth]{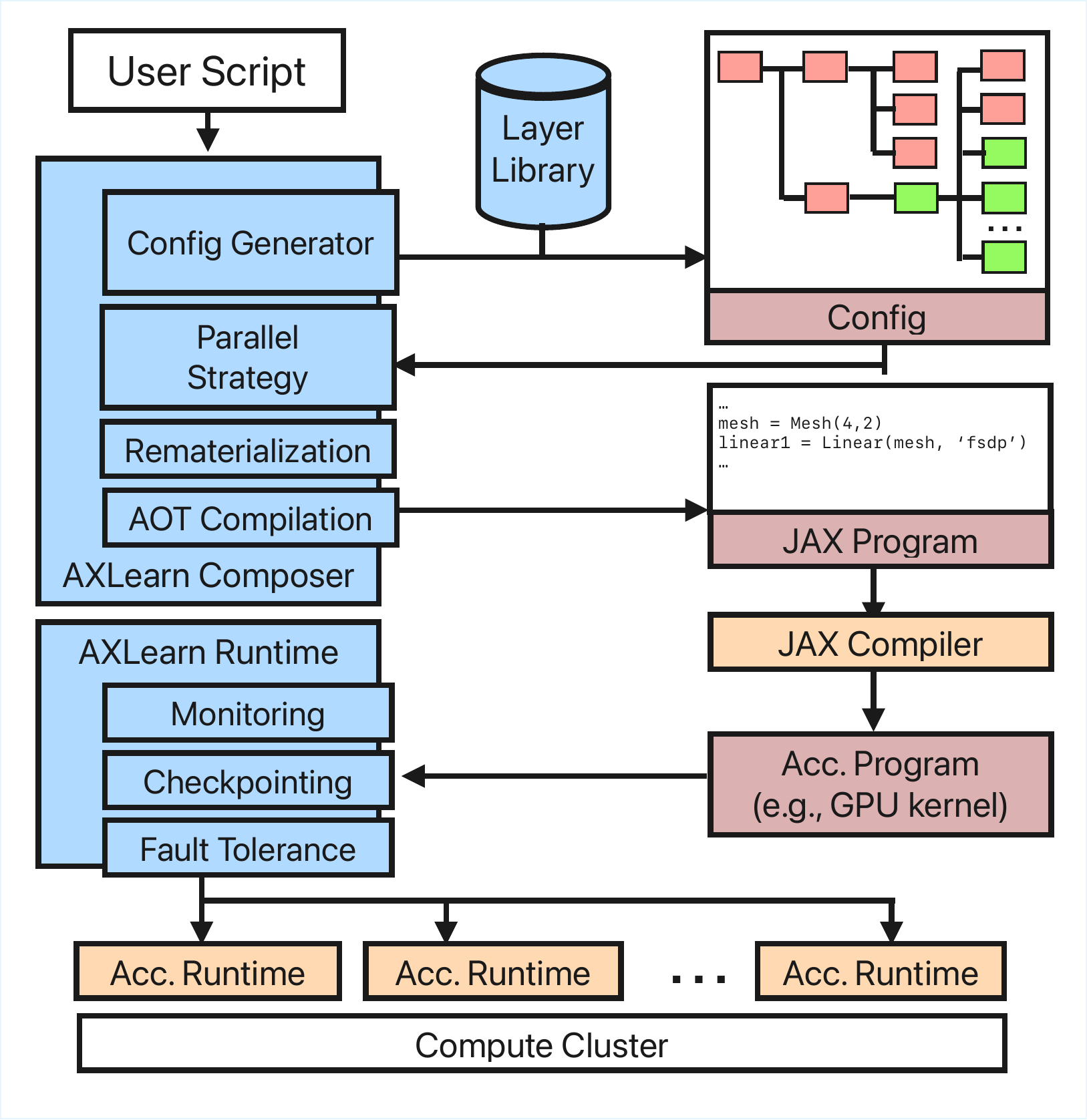}
\vspace{-3mm}
\caption{\sys's system diagram. The blue components belong to \sys.}
\label{fig:system}
\vspace{-5mm}
\end{figure}

\autoref{fig:system} shows \sys's system diagram and workflow. \sys has two key components: (1) \sys composer and (2) \sys runtime. 
A user typically uses the layer library in \sys to define a training configuration. Given such a script, the \sys composer first materializes a full JAX program, which includes selecting the appropriate mesh shape for the desired accelerator instance, applying sharding annotations for certain layers, performing auto-tuning of XLA compilation options for the target hardware, selecting appropriate attention kernels for the backend, and applying appropriate rematerialization strategies based on tagged points in the module hierarchy.
These annotations are crucial for training to run efficiently. 
The JAX program and compiler options are then passed to the XLA compiler to generate the accelerator program (including e.g. CUDA kernels),
which is then orchestrated via the \sys runtime on distributed hardware (e.g., Kubernetes) using accelerator-specific runtime (e.g., CUDA runtime). \sys runtime monitors the execution of the accelerator program and provides additional capabilities like efficient checkpointing, monitoring, and fault tolerance.
\section{\sys Composer}
\label{sec:compiler}


\subsection{Modular Configuration}

\parab{Hierarchical Configuration} TorchTitan, DeepSpeed, Flax, and most other libraries utilize ``flat'' config layouts that provide a bird's-eye view of all configurations. However, a flat layout can become unwieldy as the number of layers grows, or when switching implementations of layers (in which case, one must flatten multiple combinations of configurations). In \sys, users instead specify training configurations via composition, which logically forms a tree.
As an example, a Transformer layer's configuration consists of multiple child layer configurations:

\begin{minted}{python}
class TransformerLayer(Module):
  class Config(Module.Config):
    self_attention: AttentionLayer.Config
    feed_forward: FeedForwardLayer.Config
    ...
\end{minted}

Child configurations are encapsulated and can be specified independently of the parent: \texttt{TransformerLayer}'s config does not directly specify the hyperparameters of its child layers, allowing users to switch between different implementations of child layers.
For instance, one can decide to replace \texttt{feed\_forward} with MoE without changing the config of \texttt{TransformerLayer}.
Further, since layers can be reused between models, the configs are often initially partially specified. 

Note that we have not configured an input dimension on \texttt{feed\_forward}; instead, \texttt{hidden\_dim} is configured to be a function of the (as of yet unspecified) input dimension. This allows the parent \texttt{TransformerLayer} to set \texttt{input\_dim} when the layer is instantiated:

\begin{minted}{python}
class TransformerLayer(Module):
  def __init__(self, cfg):
    # feed_forward uses the same input_dim.
    cfg.feed_forward.set(
      input_dim=cfg.input_dim)
    self._add_child("feed_forward",
      cfg.feed_forward)
\end{minted}

By partially defining configs and propagating from parent to child, configurations often only need to be specified once, commonly at the layers near the root of the tree.
So long as the parent and child agree on a configuration interface (often just input and output dimensions), arbitrarily complex configs can be constructed while keeping layers modular. 

\parab{Config traversal}
Because configs are hierarchical and encapsulated, one can apply arbitrary modifications by traversing the config tree. We call such traversal a ``config modifier''.
The illustrate, the following snippet recursively replaces any target config with \texttt{new\_cfg}:

\begin{minted}{python}
def replace_config(cfg, tgt, new_cfg):
  def enter_fn(child):
    for key, value in child.items():
      if isinstance(value, tgt.Config):
        new_cfg.set(**value.items())
        child.set(key, new_cfg)
  cfg.visit(enter_fn=enter_fn)
\end{minted}

This can be used to apply MoE in the following manner:
\begin{minted}{python}
# Replace any FFN with MoE.
replace_config(
  trainer_cfg, 
  target=FeedForwardLayer,
  new_cfg=MoELayer.
    default_config().set(...),
)
\end{minted}

Indeed, this roughly 10-line code snippet is used to apply MoE to over 1,000 experiment configs, without any additional changes to other modules.




\subsection{Generating JAX Programs}


\parab{Config-based parallelism}
\sys natively integrates parallelism support in every relevant layer for all common parallelism strategies, including fully-sharded data parallelism~\cite{Rajbhandari2019zero}, pipeline parallelism~\cite{huang2019gpipe}, expert-parallelism~\cite{moe_dense_speed}, sequence parallelism~\cite{li2023sequence}, and tensor model parallelism~\cite{narayanan2021megatron}. 
This means that users do not need to implement parallelism directly, but instead specify their desired parallelism strategy via configuration. This contrasts with Flax or PyTorch where sharding is not a native concept in the layer library, and thus code changes may be necessary depending on parallelism strategy. Users also have granular control over how specific parameters in specific layers are partitioned.

\begin{figure*}[t]
\centering
\includegraphics[width=0.75\linewidth]{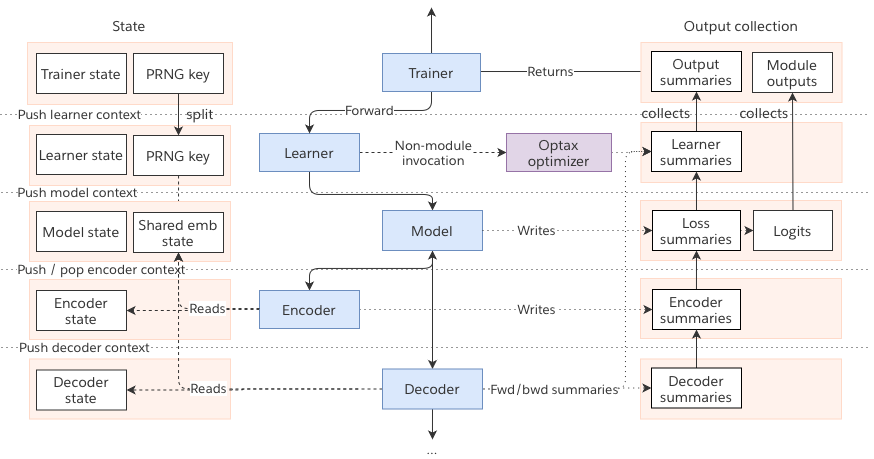}
\caption{Invocation Context. Module invocations push contexts to the stack, which retrieve child states, split PRNG keys, and create child output collections. Upon returning, contexts are popped, collecting outputs into the parent collection. The context stack can be programmatically traversed to retrieve shared state, allowing features like tied weights to preserve encapsulation.}
\label{fig:context}
\end{figure*}

\parab{Memory optimizations}
\sys has several built-in mechanisms for optimizing high-bandwidth-memory (HBM) usage which are necessary for large-scale training.
First, \sys layers natively support configuring rematerialization (a.k.a., activation checkpointing or remat). Depending on the hardware, the remat strategy must be tuned to selectively save or recompute certain activations in the backward pass. In \sys, common remat points (such as the attention QKV projections and output) are ``tagged'' with names, such that users can selectively target remat points to decide which activations to save in accelerator memory, offload to CPU memory (if supported by hardware), or recompute. Users can also employ programmatic remat strategies, such as only saving the output of linear layers.
Second, \sys implements optimizer state offloading to CPU memory. This is essential for training models with more than hundreds of billions of parameters on certain platforms like TPU v5e where HBM is limited, and where sharding beyond a certain point can be inefficient. 

\parab{Hardware-dependent optimizations}
In order to support efficient training, it is necessary to apply target-dependent parallelism strategies. 
\sys introduces the concept of ``mesh rules" to address this problem. 
A mesh rule describes a mapping from accelerator types to config modifiers, which allows users to express complex per-target optimizations with succinct and self-contained configs. \sys uses these rules to automatically apply the appropriate modifiers based on the target platform.
\autoref{appendix:mesh_rules} shows a roughly 10 line code snippet that configures training to use FSDP within TPU v5e slices and data-parallel across slices, offloading activations from dot products to host memory and enabling INT8 training.

While the XLA compiler can fuse most operations to produce optimized target-specific code, in some cases we can achieve better performance with custom kernels. Again owing to the modular nature of \sys layers, enabling custom kernels only requires simple configuration changes. For instance, \sys provides a FlashAttention~\cite{dao2022flashattention, dao2023flashattention2} layer, which can be used as a drop-in replacement for the default attention layer. Like above, this can be expressed as a config modifier in a mesh rule.
Behind the scenes, the FlashAttention layer transparently dispatches kernels based on the backend: on GPU, cuDNN~\cite{cuDNN} is used when possible, falling back to a custom Pallas~\cite{pallas} kernel for cases like block-sparse attention where cuDNN is not supported; on AWS Trainium, the Nki kernel from AWS Neuron Toolkit~\cite{neuron-nki} is used; and on TPU, the SplashAttention Pallas kernel in JAX~\cite{jax2018jax} is used.

We note that efficient hardware-agnostic training is largely possible because of several core design choices in \sys. First, \sys layers are already sharding and remat aware. This allows complex strategies to be expressed with configuration rather than code. Second, all components are implemented as strictly encapsulated modules. This allows expressing optimizations like quantization as a replacement of \texttt{DotGeneral} layers~\cite{xla_operation_semantics} with their quantization-aware equivalents. Third, configurations are entirely Python-based. This allows utilizing constructs like recursion to traverse and modify configs, which would otherwise not be possible with a DSL.

\parab{AOT compilation}
A core advantage of a compilation-based approach is that many of the errors that would otherwise be encountered in a full scale distributed run can be checked entirely locally, from the convenience of a single host environment. \sys provides native support for JAX Ahead-of-Time (AOT) compilation, which allows users to analyze the memory and FLOPS utilization of a training program without executing a single line of the program, including catching errors like OOMs that would otherwise result in wasted resources.
Because the same codepath is used for AOT and actual training, users can be confident that a program that AOT-compiles will run at a larger scale.
This saves considerable time and resource costs when scaling development to large teams.

\subsection{Maintaining States across Module Boundaries.} 
In order to be transformed by JAX primitives such as \texttt{jit} and \texttt{grad}, JAX programs must be purely functional, which means that they must be stateless.
However, neural network training is inherently stateful: it is often necessary to store model parameters, maintain pseudo-random number generators (PRNGs), collect training summaries, and aggregate outputs. 
The canonical way to maintain state in a functional system is for the caller to pass in ``side inputs'' (such as layer parameters) along with the main inputs for the method. The outputs include not only the method results, but also ``side outputs'' such as summaries and state updates.
Rather than requiring the user to maintain these side inputs and outputs, \sys introduces an abstraction called the \texttt{InvocationContext}, depicted in \autoref{fig:context}.
When a parent module invokes a child module, an \texttt{InvocationContext} for the child module and state is automatically pushed to the stack, which transparently ``splits'' the PRNG key and creates a new data store for any training summaries or outputs saved by the child module.
When a child module returns, the context is popped off the stack, which transparently collects child summaries and outputs into the parent data store.
Analogously to the traditional call stack, user code does not need to be aware of the \texttt{InvocationContext}.
This allows users to implement modules in a familiar imperative fashion while retaining the functional properties demanded by JAX.

A subtle design decision is that \texttt{InvocationContext}s contain references to modules, but not vice-versa.
This allows contexts to be accessed outside of the module hierarchy, including from arbitrary function calls that may not have references to a module, allowing deep integration with 3rd party libraries that are not natively aware of the \sys state system (e.g., optax optimizers~\cite{optax5}); as well as compatibility with codepaths with unique execution behavior, such as JAX's \texttt{custom\_vjp} backward pass.

The design of \texttt{InvocationContext} is a key factor allowing complex architectures to be well-encapsulated.
In other libraries, users may be required to perform nested instance attribute access in order to access shared state between layers. However, this requires that module implementations to be intricately aware of other modules being used and where they are in the hierarchy. 
In \sys, the system layer can instead transparently traverse the \texttt{InvocationContext} hierarchy to retrieve shared states, which allows module implementations to remain completely unaware of other modules in the system.

\section{\sys Runtime}

\parab{Monitoring and profiling.} 
\sys's runtime provides observability at several layers. At the hardware layer, \sys natively integrates with JAX's profiler to provide insights into inefficiencies introduced by input pipelines, sharding strategies, compiler behavior, or other aspects of the program.
For large scale distributed training, 
\sys supports a generic measurement interface that can be used to record arbitrary events such as the start of training or the start of a step. 
These events can be used to measure end-to-end inefficiencies such as those introduced by hardware provisioning or checkpointing recovery, which can be captured via metrics like overall job goodput.

\begin{table*}[t]
    \caption{LoC analysis of deep learning systems. $N$ denotes number of modules in a system, and $M$ denotes number of variants of a feature. LoC Estimates represent changes necessary within each system's API interfaces to integrate a single variant of RoPE or MoE under a standard production setting. Note that in \sys, 0 LoC changes to existing interfaces are necessary.}
    \begin{center}
	\begin{small}
		\centering
		\begin{tabular}{@{}c|c|c|c|c@{}}
		\toprule
        System & LoC-Complexity(RoPE) & LoC-Complexity(MoE) & LoC Estimate (RoPE) & LoC Estimate (MoE) \\
	    \hline
        Megatron-LM & $O(NM)$ & $O(N)$ & 400 & 20 \\
        \hline
        DeepSpeed & $O(NM)$ & $O(NM)$ & 320 & 4000 \\
        \hline
        TorchTitan & $O(NM)$ & $O(NM)$ & 240 & 400 \\
        \hline
        Flax & $O(NM)$ & N/A & 600 & N/A \\
        \hline
        Praxis & $O(NM)$ & $O(M)$ & 300 & 5 \\
        \hline
        MaxText & $O(NM)$ & $O(NM)$ & 200 & 300 \\
        \hline
		\textbf{\sys} & $O(1)$ & $O(1)$ & 0 & 0 \\
		\bottomrule
		\end{tabular}
	\end{small}
    \end{center}
    \label{tab:loc-complexity}
\end{table*}

\parab{Checkpointing.} \sys's default checkpointing capabilities are similar to that of \verb|orbax|~\cite{Orbax}, but supports multiple cloud storage backends, including AWS S3 and Google Cloud Storage (GCS), and provides additional memory optimizations for large-scale checkpointing.
Checkpoints are saved asynchronously during training to avoid bottlenecks on slow networks, blocking only in rare cases where the checkpointer is waiting on a prior serialization to complete. A gargage collector runs in the background to remove old checkpoints according to a user-configurable policy to reduce storage costs. 

\parab{Failure detection.} 
Failures are inevitable for large-scale training spanning thousands of chips. 
A core problem is therefore identifying when and where a failure happens.
To facilitate detection, \sys provides several components. First, \sys runtime has a configurable ``watchdog'' that monitors the step time and hardware utilization of a host. Upon observing low hardware utilization or abnormal step times, the watchdog can be configured to force a restart of the host, alert an on-call for manual intervention, or dump stack traces for debugging. 


\parab{Failure recovery.}
Efficient recovery is critical to ensure high hardware utilization and goodput. 
Large-scale training usually involves multiple data-parallel replicas with replicated weights. 
Upon a failure of a data-parallel replica, the checkpoint can be restored directly from a healthy replica and broadcasted through the fast interconnect to all other healthy or newly provisioned replicas.
Applying a similar strategy, persistent compilation caches can be configured to eliminate any startup compilation time, as compilation artifacts can be entirely reused across restarts of the same model.
Finally, \sys builds on native Kubernetes features to support slice-level hot-swap. In this case, the \sys scheduler over-provisions spare replicas within the same cluster, allowing failed nodes in an ongoing training job to be rapidly substituted with healthy nodes. Meanwhile, the over-provisioned hardware can still run low-priority jobs to reduce resource waste, or be sent for inspection and repair. 

\section{Unifying Training and Inference}
One surprising discovery during the process of building \sys is that we can get an efficient inference engine by reusing a substantial subset of \sys components. 
Though inference performance is not our design goal, \sys achieves significantly higher performance than a state-of-the-art inference engine, vLLM~\cite{kwon2023pagedattention}, on TPUs. (See \autoref{sec:evaluation} for details.)

\sys's design allows modules to easily be adapted to optimize for decoding. For example, because the attention layer's KV cache is an encapsulated component, an attention layer can incorporate inference-friendly cache layouts without changing the attention layer.
This allows incorporating new inference techniques like continuous batching~\cite{yu2022orca}, disaggregated prefill and decode~\cite{zhong2024distserve}, and paged KV cache~\cite{kwon2023pagedattention} without re-implementing models and layers. While we currently support TPU, we believe with additional effort we can support unified training and inference on other backends.

\begin{table*}[t]
\caption{Comparing \sys with state-of-art systems on training performance. Systems within each hardware group are benchmarked on identical hardware.}
\begin{center}
\begin{small}
\begin{tabular}{c|c|c|c|c|c}
\toprule
Model & Hardware & System & Iteration Time (s) & MFU & Throughput (tokens/s)\\
\hline
\multirow{9}{*}{Llama2-7B} & \multirow{4}{*}{32 $\times$ H100-8} & PyTorch FSDP & 2.6 & 29.9\% & 1.6M  \\
\cline{3-6}
&  & Megatron-LM & 1.7 & 44.9\% & 2.5M \\
\cline{3-6}
&  & MaxText & 1.4 & 54.7\% & 3.0M \\
\cline{3-6}
&  & \textbf{\sys} & 1.4 & 54.2\% & 3.0M \\
\cline{2-6}
& \multirow{3}{*}{tpu-v5p-512} & PyTorch XLA FSDP & 3.5 & 46.7\% & 1.2M \\
\cline{3-6}
&  & MaxText & 2.7 & 61.6\% & 1.6M \\
\cline{3-6}
&  & \textbf{\sys} & 2.5 & 66.2\% & 1.7M \\
\cline{2-6}
& 64 $\times$ Trainium2-16 & \textbf{\sys} & 1.2 & 24.2\% & 3.5M \\
\hline
\multirow{9}{*}{Llama2-70B} & \multirow{4}{*}{64 $\times$ H100-8} & PyTorch FSDP & 10.6 & 34.7\% & 396K  \\
\cline{3-6}
&  & Megatron-LM & 7.8 & 47.2\% & 538K \\
\cline{3-6}
&  & MaxText & 9.4 & 39.1\% & 446K \\
\cline{3-6}
&  & \textbf{\sys} & 9.2 & 40.0\% & 456K \\
\cline{2-6}
& \multirow{3}{*}{tpu-v5p-1024} & PyTorch XLA FSDP & \multicolumn{3}{c}{OOM}  \\
\cline{3-6}
&  & MaxText & 12.3 & 64.4\% & 341K \\
\cline{3-6}
&  & \textbf{\sys} & 11.6 & 68.0\% & 360K \\
\cline{2-6}
& 64 $\times$ Trainium2-16 & \textbf{\sys} & 11.2 & 25.0\% & 374K \\
\hline
\multirow{4}{*}{Qwen-3 30B-A3B} & \multirow{2}{*}{tpu-v5p-1024} & MaxText & 12.97 & 31.31\% & 1.3M \\
\cline{3-6}
&  & \textbf{\sys} & 12.86 & 31.58\% & 1.3M \\
\cline{2-6}
& \multirow{2}{*}{64 $\times$ B200-8} & Megatron-LM & 4.10 & 20.20\% & 4.1M \\
\cline{3-6}
&  & \textbf{\sys} & 4.31 & 19.22\% & 3.9M \\

\bottomrule
\end{tabular}
\end{small}
\end{center}
\label{tab:training}
\end{table*}

\section{Evaluation}
\label{sec:evaluation}


\subsection{Modularity of Configurations}
We analyze modularity using LoC-Complexity, motivated at length in \autoref{sec:background}. Specifically, we analyze the asymptotic LoC changes required to re-parameterize each system to support RoPE and MoE.
When measuring extensibility, we focus on LoC changes incurred in \textit{existing} modules in the system as opposed to the new functionality itself. This allows for a comparable analysis, and in practice, the same feature across systems tends to incur similar LoC up to some constant.

\autoref{tab:loc-complexity} shows the summary of LoC-Complexities across systems. We omit PyTorch FSDP, which is not an LLM library; Haiku, which has no implementation of RoPE or MoE; and Pax, is replaced with its layer library Praxis.
For a concrete evaluation, we also provide LoC estimates based on the changes required in a production codebase. While many of the listed systems provide a few sample implementations, a production codebase can be orders of magnitude larger. 
We cannot disclose specific details of the scale of our internal model architectures and hyper-parameters. 
Therefore, we propose estimates under a realistic setting of a codebase with 20 model variants (e.g., variants of a GPT model) and 10 variants of attention (i.e., kernels, kv cache strategies, etc.). 
\textbf{Notation:} Let $N$ be the number of modules in a system, and $M$ be the number of variants of the feature to be added. For example, MoE has variants of routing or gating mechanisms.

\parab{\sys (ours)}
In \sys, RoPE and MoE are strictly encapsulated. \autoref{sec:compiler} provides a 10-line code snippet to integrate both features into any experiment config. Internally we use such a code snippet to configure over 1,000 experiments to use RoPE, MoE, or both. As we scale the number of modules or RoPE or MoE variants, we require no changes to any existing interfaces, achieving constant LoC-Complexity.

\parab{Other Systems} All other systems we have studied require  $O(NM)$ to incorporate RoPE and MoE, with the following exceptions. Like \sys, Megatron-LM defines a MoE layer in place of the standard MLP. However, each \text{MLP} introduces MoE-specific fields in its signature, so that any module composing a linear submodule incurs at least one LoC change for MoE-specific arguments. This scales with $O(N)$. Like \sys, Praxis uses a ``template'' composition approach, which allows LoC-Complexity(MoE) to be $O(M)$: each MoE variant incurs a small 5 LoC change. Flax has no public example of MoE so we exclude it from our analysis. See \autoref{appendix:loc-analysis} for details on the LOC analysis.

\subsection{Performance on Different Hardware}
We compare \sys training performance with PyTorch FSDP, Megatron-LM, and MaxText, which achieve state-of-the-art performance on GPU and TPU. All comparisons are conducted on identical hardware to ensure fair evaluation---systems sharing a hardware group in \autoref{tab:training} are benchmarked on the same cluster with the same number of accelerators.
We evaluate both dense and MoE models: Llama2 7B and 70B (dense), and Qwen-3 30B-A3B (MoE), on four hardware backends: (1) 256/512 H100 GPUs (32/64 AWS P5d instances, each with 8 H100 GPUs); (2) 512 B200 GPUs (64 AWS P6 instances, each with 8 B200 GPUs); (3) TPU-v5p-512/1024 (64/128 GCP Cloud TPU hosts, each with 4 chips); and (4) 1024 Trainium2 (64 AWS trn2 instances, each with 16 Trainium2 chips). All runs use a global batch size of 1024.
Note that since Pytorch FSDP does not run on TPU, we use Pytorch XLA FSDP as its closest replacement~\cite{pytorch_xla_fsdp}. Megatron-LM is a GPU-only system, so we only report results on GPU. None of the baselines support Trainium, so we only report results for \sys.

\begin{figure}[t]
    \begin{subfigure}{0.48\linewidth}
    \includegraphics[width=\linewidth]{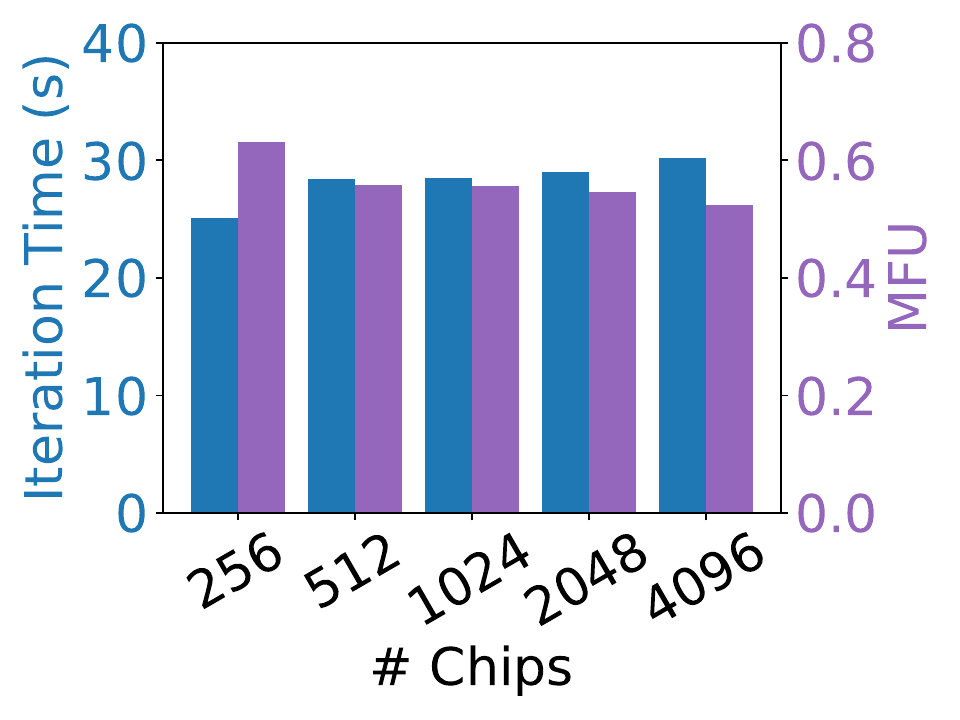}
    \caption{Model A (70B)}
    \end{subfigure} \hfil
    \begin{subfigure}{0.48\linewidth}
    \includegraphics[width=\linewidth]{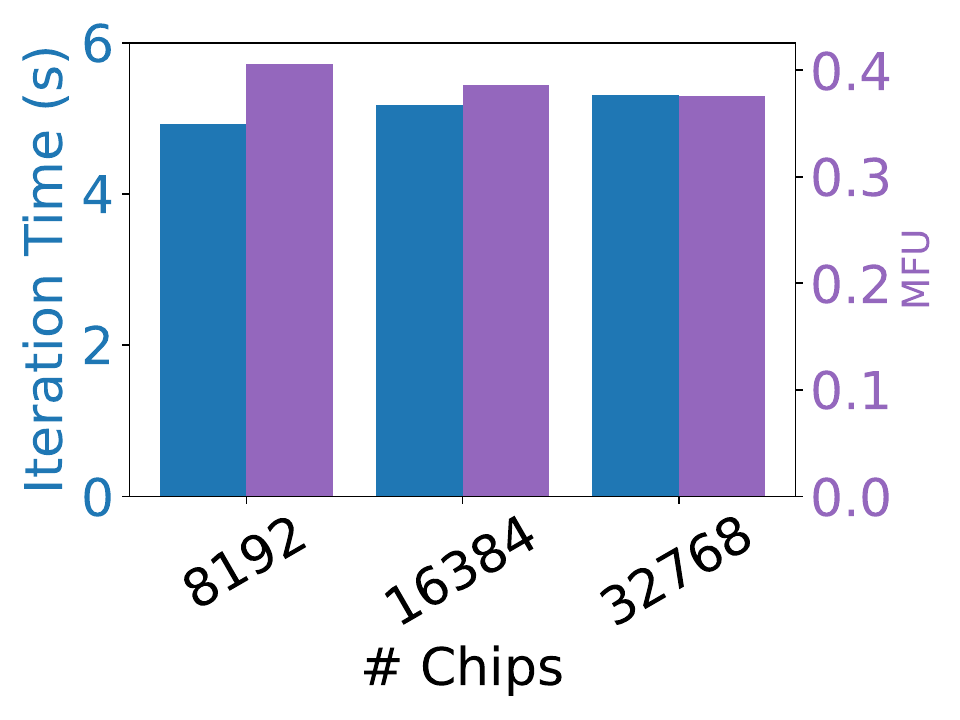}
    \caption{Model B (150B)}
    \end{subfigure}
\caption{\sys's training throughput and MFU when scaling the number of TPUs.}
\label{fig:scalability}
\end{figure}

\autoref{tab:training} summarizes the performance results. On TPUs, \sys achieves state-of-the-art performance, where MaxText lags slightly behind, likely due to its choices of rematerialization.
On the other hand, PyTorch XLA FSDP fails to run entirely with out-of-memory errors.
On H100 GPUs, \sys outperforms PyTorch FSDP due to fine-grained control over activation checkpointing and better support for memory bound operations such as RMSNorm and RoPE. Megatron-LM has stronger performance on H100 GPUs over \sys because PyTorch currently have finer-grained scheduling capability over XLA. However, using XLA allows \sys to be hardware-agnostic. This is a trade-off we are willing to take.



To test \sys's scalability, we perform a weak-scaling study of two production models. Model A is a 70B model with 4,096 context length, and Model B is a 150B model with 8,192 context length. Fixing the per-device batch size, for Model A, scaling from 256 to 4,096 chips reduces MFU from 63.0\% to 52.4\%; and for Model B, scaling from 8,192 to 32,768 chips reduces MFU from 40.6\% to 37.6\%. These numbers demonstrate that \sys achieves very close to linear scaling, shown in \autoref{fig:scalability}. We note that MFU for the 150B model is lower due to the need to limit the global batch size at 32,768 chip scale for good training convergence. All scaling experiments for the 150B model has 1/16 per chip sequence length compared to 70B experiments.

\begin{table}[t]
\begin{center}
\begin{small}
\caption{Comparing \sys with vLLM on inference performance for Llama 2 models on Time-To-Fist-Token (TTFT), Time-Per-Output-Token (TPOT), and throughput on TPUs.}
\begin{tabular}{c|c|c|c|c}
\toprule
Model & System & TTFT & TPOT & Throughput\\
\hline
\multirow{2}{*}{7B} & vLLM & 538.6ms & 22.4ms & 1117 tokens/s\\
\cline{2-5}
& \textbf{\sys} &  40.1ms & 9.1ms & 3125 tokens/s \\
\hline
\multirow{2}{*}{70B} & vLLM & 80s & 189.8ms & 705 tokens/s \\
\cline{2-5}
& \textbf{\sys} & 150.5ms & 28.1ms & 1139 tokens/s \\
\bottomrule
\end{tabular}
\label{tab:inference}
\end{small}
\end{center}
\end{table}

A key benefit of \sys's modular design is that the same training codebase can serve inference workloads with minimal additional effort. We compare the inference performance of Llama2 7B and 70B models between \sys and vLLM. We use the ShareGPT dataset for the prompts.
For the 7B model, the benchmark is performed on a Google Cloud TPU v5p-8 VM instance, with maximum input length of 1024 and maximum output length of 256. For the 70B model, the benchmark is on a Google Cloud TPU v6e-8 VM instance\footnote{vLLM does not support the 70B model on v5p-8 VM by the time of benchmark.}, with maximum input length of 1,800 and maximum output length of 256. As \autoref{tab:inference} shows, \sys achieves 500x and 6x speedup in TTFT and TPOT over vLLM, respectively. In terms of throughput, \sys is 2.8x faster for 7B model inference and 1.6x faster for 70B model inference. We note that TPU support for vLLM is still experimental, which likely contributes to the performance gap. Our goal is not to position \sys as a specialized inference engine, but to demonstrate that a modular training framework can achieve production-grade inference performance with minimal additional effort.


\subsection{Failure Recovery Latency}
\begin{figure}[t]
\centering
\includegraphics[width=0.8\linewidth]{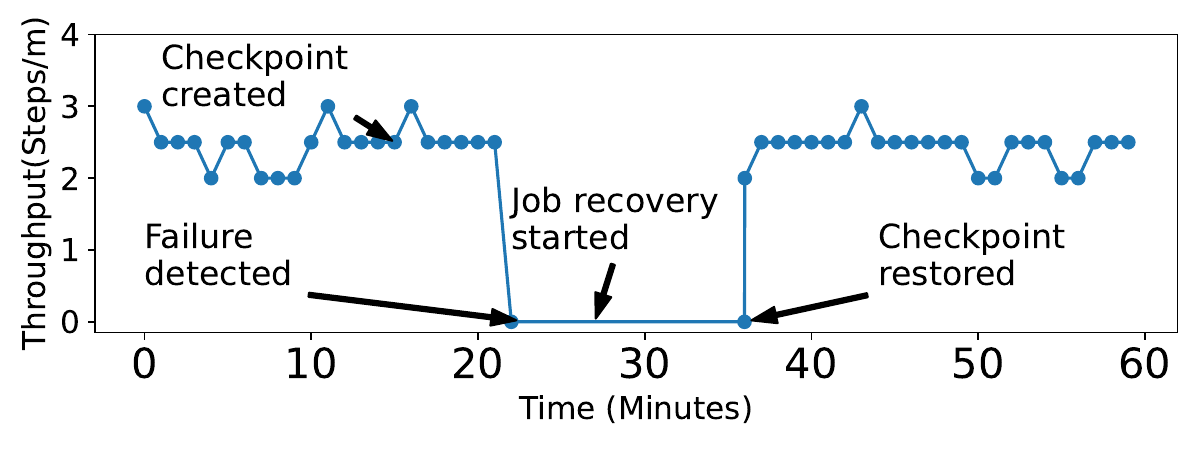}
\caption{\sys's recovery latency under a hardware failure.}
\label{fig:failover}
\end{figure}

We measure the speed of recovering from a hardware failure via analyzing a production training job. \autoref{fig:failover} show training throughput over time on 32,768 TPUs over a 1-hour window that contains a complete event which the job recovers from a hardware failure. Since checkpointing is performed asynchronously, no training throughput reduction is observed during checkpoint creation. When hardware fails, \sys detects the training job has failed immediately and initiates slice-level hot-swap, completing the hot-swap within 4 minutes. Checkpoint restoration finishes in 9 minutes after the hot-swap completes. In total, the failure results in 21 minutes of lost training time, which includes both the downtime during hot-swap and checkpoint restoration, as well as the lost progress from training step completed after the most recent checkpoint before the failure occurred.

\subsection{Experience of using \sys at Apple}
\label{sec:experience}

\sys began development in late 2021 using PyTorch with a small team of engineers. At the time, signs of the Transformer architecture becoming a de facto choice were evident (e.g., BERT~\cite{Devlin2019BERT}, T5~\cite{raffel2020t5}, GPT3~\cite{brown2020languagemodelsfewshotlearners}). Architectural convergence meant that most layers in many new models can be reused if implemented modularly. However, due to limited support for automatic parallelization, layer implementations ultimately could not be fully modularized---in many cases, different parallelization strategies meant that logic had to be rewritten, with inevitable interactions across modules.

GSPMD~\cite{xu2021gspmd} (published late 2021) meant that communication can instead be injected by a compiler. We chose to deeply integrate with XLA as a bet on the compiler-first approach. First, layers can be modularized as sharding propagation across the graph can be automatically and transparently handled. Second, we believed eventual adoption would push hardware providers to optimize compilers on their platforms, which means performance improvements can be achieved without a single LoC. As it turns out, while JAX/XLA did not initially support GPUs, today it achieves competitive performance across AWS and GCP; and is now natively supported by platforms like Trainium2. Where compilers fall short, we rely on custom kernels to close the performance gap. While \sys's modular design makes integrating custom kernels straightforward---as drop-in replacements via config modifiers---achieving peak hardware utilization on each backend requires backend-specific kernel implementations, which demands specialized expertise and ongoing maintenance as hardware evolves.

Adopting JAX/XLA meant several paradigm shifts. 
First, it shifts from imperative to functional programming, which posed a challenge as familiar patterns like mutable state were prohibited. Instead, one manually specified parameters as inputs to each module, and bubbled up outputs through the call stack. This hurt modularity as it required implementations to be intricately aware of state structure, and hurt usability as users lost the comfort of the PyTorch imperative style. This motivated the design of \texttt{InvocationContext}, which allowed state to be decoupled, outputs to be collected across the module hierarchy, and layer implementations to be implemented in familiar PyTorch fashion.
Second, it shifts focus to the cloud. JAX/XLA initially only supported Google Cloud TPU, which meant that a number of internal components could not be directly used. 
As GCP was in a nascent state with limited TPU capacity, we knew we could not rely on it entirely, and hoped to leverage our internal compute infrastructure. To that end, we designed the system to allow all components of training to be configurable. For example, we initially leveraged Flax's GCS-compatible checkpointer to bootstrap the layer library, which we later replaced to support internal storage backends. This migration was completely seamless with no changes to user code, which was only possible due to \sys's modular design. 

As we scaled development, we faced additional barriers. 
First, resource contention was crucial due to limited TPU capacity within GCP. Aside from leveraging other backends, we realized that a significant portion of resources were consumed by jobs with low resource utilization or no progress due to preventable errors. However, XLA compilation inherently operates on device abstractions---many errors (e.g., OOMs due to suboptimal sharding) can be theoretically caught from a local machine. We therefore capitalized on JAX's compiler-first approach by deeply integrating AOT-compilation, which allowed users to debug training entirely on CPU. This has allowed development to scale even with limited capacity.
Second, we learned the hard way that ML testing practices are often insufficient. While it's common to unit test layers, ML experiments heavily depend on correctness of configs. 
With a large codebase, changes to one experiment can inadvertently affect others, which are difficult to catch with unit or integration testing (e.g., subtle changes in training dynamics). 
To address this issue, \sys introduces the notion of ``golden configuration'' tests: key training configs are serialized into human readable format and committed along with code changes. This allows changes to produce consistent and reviewable diffs, to trigger relevant code-owner reviews, avoid code-contribution-level conflicts, and provide a traceable commit history of experiments. 
Lastly, public cloud infrastructure can fail in opaque ways. In contrast to an internally managed cluster, we learned that we must account for opaque failures often out of our control, including hardware failures, ICI failures, silent data corruptions, kernel panics, file system throttling, and more. While we have built many layers of resiliency, debugging and mitigating many of these failures required close collaboration with Google, Amazon, and Nvidia.

Today, \sys has grown from a handful of developers training models at million parameter scales, to hundreds of developers training models at billion-to-trillion parameter scales. It actively supports over 10,000 experiments under development at a given time, running across tens of different hardware clusters. Some of the models trained with \sys now power features used by over a billion users, including intelligent assistants, multimodal understanding and generation, and code intelligence.
\section{Related Work}
\parab{Training systems for large models.}
High-performance large model training is an active research area. FSDP~\cite{zhao2023pytorchfsdp} and DeepSpeed~\cite{Rajbhandari2019zero} use sharding to reduce GPU memory usage.
Megatron-LM~\cite{narayanan2021megatron} combines data, tensor-model, and pipeline parallelism to train LLMs efficiently on GPUs.
MegaScale~\cite{jiang2024megascale} further scales LLM training to more than 10,000 GPUs and deals with straggler and failures in GPU clusters.
\sys's focus is complementary to these systems, with the goal of enabling engineers to use these techniques with minimal effort.

\parab{Software modularity.} 
Maintaining a modular structure is a standard software engineering practice in any complex computer system, facilitating maintenance and integration of new features~\cite{bershard1995spin, kohler2000click}. Modularity is especially important for a fast-moving machine learning system: 
at Apple, we iterate on new model architectures, hyper-parameters, and training methods every day. 
To facilitate this development, configurations must be separated from system details like parallelism or rematerialization in the simplest way possible, so that researchers can focus on model iteration.

\parab{Configuration systems for distributed workloads.} 
One of our contributions is the ability to configure distributed deep learning execution in a modular and automated manner. Configuration for distributed systems is a topic beyond deep learning. Meta's configuration system~\cite{tang2015metaconfig} can enable easy A/B testing, similar to how to we conduct A/B testing on model accuracy and training performance by comparing \sys's configurations. However, our main design goal for the configuration system is to enable modules to be re-parametrized and composed to easily integrate with new features (e.g., RoPE, MoE).

\section{Conclusion}
We have developed \sys, a modular, hardware-agnostic large model training system. \sys maintains constant complexity as the number of modules scales.
Further, \sys has comparable performance compared to other state-of-the-art training systems. Finally, we shared our development and operation experiences with \sys.

\section*{Acknowledgments}
We thank Yi Wang, Ke Ye, Dong Yin, Muyang Yu, Yi Zhang, and other key members of the Apple Foundation Model team, for their deep collaboration and valuable feedback in the development of AXLearn. We also thank Benoit Dupin and Daphne Luong for their leadership support.

\bibliography{reference}
\bibliographystyle{mlsys2025}

\appendix
\section{Mesh Rules}\label{appendix:mesh_rules}

In \sys, users can specify per-target platform configuration changes using ``mesh rules''. These rules are mappings from instance type regular expressions to config modifiers discussed in \autoref{sec:compiler}. 
In the example below, when launching training on TPU v5e, we configure training to use FSDP within-slices and data-parallel across slices, offloading activations from dot products to host memory and enabling INT8 training.
On the other hand, when launching training on H100s, we instead switch to 8-way tensor parallelism within node and FSDP across nodes, saving query, key, value, and output (QKVO) projections to HBM, as well as enabling FP8 training with delayed scaling.

These configs are all that are necessary to apply per-target optimizations, allowing users to scale training on different platforms with ease.

\begin{minted}{python}
[("tpu-v5e-256-*",
  [MeshShapeModifier.default_config().set(
      mesh_shape=mesh(data=-1, fsdp=256)),
    RematSpecModifier.default_config().set(
      remat_policies={
        "model.decoder.transformer.layer":
          RematSpec(policy=offload_dots)}),
    INT8ConfigModifier.default_config()]),
("gpu-H100-*",
  [MeshShapeModifier.default_config().set(
      mesh_shape=mesh(fsdp=-1, model=8)),
    RematSpecModifier.default_config().set(
      remat_policies={
        "model.decoder.transformer.layer":
          RematSpec(policy=save_qkvoflash)}),
    FP8ConfigModifier.default_config().set(
      fp8_amax_history_length=128)])]
\end{minted}

\section{LoC Analysis}
\label{appendix:loc-analysis}

We provide additional details and rationale of how we derived the LoC estimates in \autoref{tab:loc-complexity}.

\parab{Megatron-LM}
Megatron-LM's Transformer implementation composes \texttt{TransformerBlockSubmodules}.
However, the RoPE-specific parameters are flattened in the \texttt{init} signature of each model implementation, e.g. \texttt{rotary\_percent}, \texttt{rotary\_base}, \texttt{rotary\_scaling}, and \texttt{position\_embedding\hspace{0pt}\_type} in \texttt{GPTModel}.
Additionally, these parameters are propagated to submodules like \texttt{TransformerBlock}, \texttt{Transformer} \texttt{Layer}, or \texttt{Attention}.

This means to integrate a RoPE variant, one potentially incurs LoC changes to each module in each model implementation, as at minimum the RoPE parameters must be propagated down an arbitrary number of modules to the \texttt{Attention} layer.

Additionally, each model's \texttt{init} implementation includes branching logic to instantiate the RoPE embedding layer variant depending on the desired \texttt{position\_embedding\_type}. For example, \texttt{RotaryEmbedding} should be instantiated if the embedding type is ``rope'', while \texttt{MultimodalRotaryEmbedd\hspace{0pt}ing} should be instantiated when embedding type is ``mrope''.
Therefore, if we additionally consider RoPE variants, the LoC-complexity scales quadratically to $O(NM)$, as in the worst case each module must account for each variant in its \texttt{init} signature if it receives an embedding type.

To integrate MoE, Megatron-LM is able to leverage composition via \texttt{TransformerBlockSubmodules} to specify a MoE layer in place of the standard \texttt{MLP}. However, once again the encapsulation is not applied strictly. Each \text{MLP} layer implementation introduces an \text{is\_expert} field in its \text{init} signature, which is propagated to the linear submodules. Any module that uses a linear submodule therefore needs to incur at least one LoC change, which scales with $O(N)$. Indeed, Megatron-LM itself has a number of modules that are impacted, including \texttt{ColumnParallelLinear}, \texttt{RowParallelLinear}, \texttt{Att\hspace{0pt}ention}, \texttt{CrossAttention}, and more.

If we assume the production setting of 20 model variants, with each conservatively incurring at least 20 LoC changes to integrate RoPE\footnote{Based on the changes required in \texttt{GPTModel}.}, we incur at least 400 LoC.
For MoE, if we assume 10 MLP variants\footnote{Based on Megatron-LM's own Linear variants.}, each incurring at least 1 LoC change to integrate support for \texttt{is\_expert}, we incur at minimum 10 LoC change.
In addition, if we assume at least 10 modules in the system using a linear submodule (corresponding to each linear variant), each requiring 1 LoC change to \texttt{build\_module}, we incur additional 10 LoC.

\parab{DeepSpeed}
DeepSpeed applies the ``config flattening" methodology discussed in \autoref{sec:compiler}. RoPE configs like \texttt{rotary\_dim} and \texttt{rope\_theta} are grouped under a monolithic config class, like \texttt{DeepSpeedInferenceConfig}. 
Every model implementation reads the config and overrides the method \texttt{positional\hspace{0pt}\_embedding\_type} to indicate whether RoPE should be enabled for the model by returning a specific value of the embedding type.
With this design, we can already observe that LoC-Complexity(RoPE) must be \textit{at least} $O(N)$, as each model implementation incurs LoC changes to override the necessary methods to enable RoPE\footnote{E.g., \texttt{positional\_embedding\_type} and \texttt{positional\_embedding\_config}}.

In addition, the base model implementation propagates this embedding type (and additional RoPE configs) to child layers like the self attention layer. As a consequence, every attention layer implementation, such as \texttt{DSDenseBlocked\hspace{0pt}Attention}, must first update its \texttt{init} signature to handle the input embedding type. It must also update its forward implementation to apply RoPE based on the embedding type.

In a similar way as Megatron-LM, LoC-complexity(RoPE) scales quadratically with $O(NM)$, because we must propagate embedding type and RoPE parameters down an arbitrary number of modules to the attention layer, and because each attention module in the worst case receives and must be prepared to handle all possible values of embedding type. 

If we assume conservatively 20 models with at least 6 LoC per model\footnote{To override two properties.}, we incur 120 LoC; in addition, with 10 attention variants each requiring approximately 20 LoC\footnote{Based on changes in \texttt{DSDenseBlockedAttention}.} we incur another 200 LoC.

If the MoE case, DeepSpeed requires subclassing each model from a custom subclass \texttt{DSMoETransformerModelBase}, which requires in some cases a re-implementation of most methods. For 20 model variants, each can incur 100s of LoC---in the case of DeepSpeed's QwenV2MoE, more than 200 LoC---which conservatively incurs 4,000 LoC changes.

\parab{TorchTitan}
TorchTitan adopts a similar design as DeepSpeed by using a monolithic \texttt{BaseModelArgs} subclass that flattens all configs for each model. In the case of RoPE, each such config subclass introduces RoPE specific configurations like \texttt{rope\_theta} and \texttt{rope\_scaling}. Similar to prior analysis, such a design already incurs at least $O(N)$ LoC-Complexity as each model adopting RoPE must accordingly modify its config class signature.

In addition, each model has its own \texttt{Attention} implementation. For example, the \texttt{deepseek\_v3} \texttt{Attention} implementation conditions on the value of RoPE configs (e.g. \texttt{rope\_scaling}) to decide which child RoPE layer to instantiate. In the worst case, each RoPE variant may incur the same conditional logic for each model implementation that intends to support it. This causes the LoC-Complexity to degrade to $O(NM)$ if we consider RoPE variants.

With 20 model variants, each incurs ~2 LoC on average to update its corresponding \texttt{ModelArgs} class to incorporate RoPE configs, as well as ~10 LoC for each attention implementation. In total, this incurs 240 LoC, although we note that this varies greatly between TorchTitan model implementations\footnote{Based on DeepSeekV3 and Llama4 implementations.}.

In the MoE case, TorchTitan conditionally instantiates either a MoE child layer or a standard \texttt{FeedForward} layer, based on the config field \texttt{moe\_enabled}. Because TorchTitan implements customized layers for each model (e.g., \texttt{Transfor\hspace{0pt}merBlock} for Llama, \texttt{DecoderLayer} for DeepSeekV3), this results in $O(NM)$ complexity as we must modify the corresponding transformer layer for each model for each MoE variant.

Like with RoPE, we incur at least 10 LoC for each \texttt{ModelArgs} (due to more complex configuration of MoE), as well as ~10 LoC for each attention layer implementation\footnote{Based on \texttt{TransformerBlock} and \texttt{DecoderLayer}.}. Over 20 model variants, this results in 400 LoC changes.

\parab{Flax}
Flax implements the Gemma model via mostly self-contained Transformer modules which are not shared with other implementations. To integrate RoPE, the Gemma \texttt{Trans\hspace{0pt}formerConfig} is modified to add the RoPE parameters; the Gemma \texttt{Transformer} module is modified to propagate the RoPE parameters from the config to its \texttt{Block} submodules; each \texttt{Block} submodule is modified so that its \texttt{init} signature can accommodate additional RoPE parameters; and each \texttt{Attention} module must take these RoPE parameters to finally implement the RoPE logic.
It is easy to see that the LoC-Complexity(RoPE) in this case is $O(NM)$, as for each RoPE variant, one would need to correspondingly update RoPE parameters in the top-level \texttt{TransformerConfig}, and then propagate those configs down an arbitrary number of modules to the \texttt{Attention} layer implementation.
For a single variant, the Gemma model incurs at least 30 LoC across \texttt{TransformerConfig}, \texttt{Transformer}, \texttt{Block}, and \texttt{Attention} parameterization modifications, not including the actual RoPE implementations. For 20 model variants, this results in at least 600 LoC changes.

\parab{Praxis}
Praxis is the layer library of Pax. It internally uses fiddle, a config system similar to the one in \sys, which allows it to express certain re-parameterizations using composition.
For example, Praxis uses a ``template'' approach to configure the MoE layer in each transformer layer stack. This ``template'' is configured along with several MoE configs (like \texttt{num\_experts}) in the \texttt{StackedTransformerLayer} definition, which in theory allows the integration of MoE to scale with $O(M)$, or the number of MoE variants.
Because some MoE configs are still flattened, although not to the extent of causing quadratic interactions, each MoE variant incurs ~5 LoC change\footnote{Based on the number of flattened configs.}.

However, using a composable config system doesn't guarantee strict encapsulation. 
In particular, Praxis flattens a number of RoPE-specific configs (e.g., \texttt{use\_rotary\_position\hspace{0pt}\_emb}) into each attention layer implementation, which means that it incurs at least $O(N)$ LoC-Complexity for a single RoPE variant. As it turns out, Praxis attention layers compose the actual RoPE implementation itself via defining a RoPE layer ``template" \texttt{rotary\_position\_emb\_tpl}. However, because each RoPE variant may have different configuration interfaces, the flattening ultimately means that LoC-Complexity under RoPE variants is $O(NM)$, as each RoPE variant may nevertheless require modifications to each attention layer.

For 10 attention variants\footnote{Based on number of attention layers in Praxis}, each incurring approximately 30 LoC\footnote{Based on \texttt{DotProductAttention} and \texttt{MultiQueryDotProductAttention}}, this incurs at least 300 LoC.

\parab{MaxText} 
MaxText builds on top of Flax modules, and follows a similar layer design and analysis.
MaxText's \texttt{Attention} conditions on configs (like \texttt{attention\_type} and \texttt{rope\_type}) to choose the RoPE module. This has undesirable interactions between the attention and RoPE variants; in the worst case, \texttt{Attention} must account for the cross-product of attention and RoPE variants. We can already observe this with MaxText's MLA implementation; while MLA implemented as a subtype, its RoPE logic must also be handled in its parent Attention. 
For each RoPE variant and model, we can expect at least 10 LoC\footnote{Based on Attention implementation for ``llama3.1", ``yarn", etc.}, which results in 200 LoC across 20 variants.

In a similar way, MaxText's MoE implementation details are flattened into each model's decoder.
Each model implements its own \texttt{DecoderLayer}, while a monolithic \texttt{Decoder} selects a decoder layer implementation based on the name of the model in the config. With ~10 LoC per decoder\footnote{Based on changes in Decoder.}, this results in 200 LoC for 20 model variants.
MaxText's trainer also includes MoE-specific logic: each loss function uses MoE configs to apply auxiliary losses.
With 5 LoC for each loss function\footnote{Based on \texttt{loss\_fn} and \texttt{dpo\_loss\_fn}.}, across 20 model variants this results in additional 100 LoC.

\end{document}